\documentclass[journal,twoside,web]{ieeecolor}
\usepackage{generic}
\usepackage{cite}
\usepackage{amsmath,amssymb,amsfonts}
\usepackage{graphicx}
\usepackage{algorithm,algorithmic}
\usepackage{hyperref}
\usepackage{booktabs}
\usepackage{placeins}
\hypersetup{hidelinks=true}
\usepackage{textcomp}
\usepackage{multirow}
\usepackage{booktabs,makecell}
\usepackage[normalem]{ulem}
\def\BibTeX{{\rm B\kern-.05em{\sc i\kern-.025em b}\kern-.08em
    T\kern-.1667em\lower.7ex\hbox{E}\kern-.125emX}}
\begin{document}
\title{MRG-R1: Reinforcement Learning for Clinically Aligned Medical Report Generation}
\author{Pengyu Wang,  Shuchang Ye, Usman Naseem, Jinman Kim \IEEEmembership{Member, IEEE}
\thanks{Corresponding author: Jinman Kim and Usman Naseem}
\thanks{Pengyu Wang, Shuchang Ye, and Jinman Kim are with the School of Computer Science, The University of Sydney, Camperdown, NSW 2006, Australia (email: pwan0442@uni.sydney.edu.au; shuchang.ye@sydney.edu.au; jinman.kim@sydney.edu.au).}
\thanks{Usman Naseem is with School of Computing, Macquarie University, Macquarie Park, NSW 2113, Australia (email: usman.naseem@mq.edu.au).}
}

\maketitle

\begin{abstract}

Medical report generation aims to automatically produce radiology-style reports from medical images, supporting efficient and accurate clinical decision-making. 
However, existing approaches predominately rely on token-level likelihood training, which favors local lexical matching and leaves clinical correctness under-specified in the training objective. This behavior can be attributed to token-level likelihood optimization, which rewards surface-form agreement and therefore fails to directly encode constraints on medically accurate findings. 
To address this objective mismatch, we introduce a semantic-driven reinforcement learning (SRL) framework for medical report generation, named MRG-R1, which directly optimizes report-level clinical correctness rather than token-level likelihood. The key module is a clinically grounded report-level reward function, which reinforces semantic agreement in clinically relevant findings between generated and reference reports, thereby enabling learning signals that explicitly constrain medical correctness beyond surface linguistic alignment. 
Our evaluations show that the proposed framework improves the accuracy and coverage of clinically relevant findings in generated reports, and that MRG-R1 achieves state-of-the-art clinical efficacy  on the IU X-Ray and MIMIC-CXR benchmark datasets.


\end{abstract}


\section{Introduction}

Automatic medical report generation (MRG) aims to produce radiology-style narratives from medical images, documenting diagnostically relevant findings in a format familiar to radiologists. This is an important and timely task, as the growing use of medical imaging has substantially increased radiologists’ workload \cite{kwee2021workload}. Interpreting large volumes of images and translating them into accurate diagnostic reports requires considerable expertise and time. Under workload pressure, reporting remains vulnerable to errors, including missed subtle abnormalities, mis-specified uncertainty, and inconsistent terminology \cite{kasalak2023work,peng2018negbio}. In real clinical workflows, these challenges are further compounded by heterogeneous inputs such as variable acquisition protocols, image quality, and the presence of lines and devices \cite{acosta2024impact}. Against this backdrop, MRG has emerged as a promising approach to improve reporting efficiency, reliability, and clinical accuracy.

Recently, automatic MRG has garnered significant interest and achieved substantial advances with deep learning applied to healthcare \cite{ye2024dynamic,chen2020generating,chen2022cross,jing2017automatic,wang2018tienet}. However, as reported in prior work \cite{miura2021ifcc}, existing methods can generate fluent, radiologist-style reports, but fail to ensure clinical correctness. One of the key reason for this limitation is that most existing models learn with token-level objectives: they decompose reporting into autoregressively predicting the next token, which rewards surface n-gram overlap (i.e., local sequences of adjacent words). This makes it easy for models to produce sentences that are locally coherent in surface form, yet misaligned with the underlying clinical meaning. Because token-level training rewards local lexical plausibility, models may generate radiology-style statements that appear reasonable in isolation but fail to preserve globally coherent clinical semantics across the full report. Early MRG methods were subject to these limitations due to their reliance on encoder-decoder architectures with convolutional neural network (CNN) backbones for visual features and long short-term memory (LSTM) or gated recurrent unit (GRU) decoders for sentence generation \cite{jing2017automatic,li2018hybrid,wang2018tienet}. Transformers improve global context modeling and parallelization, helping capture dependencies across multiple sentences and sections and enabling stronger cross-modal alignment \cite{chen2022cross,liu2021exploring}. More recently, large vision-language models (LVLMs), which couple high-capacity visual encoders with instruction-tuned (fine-tuned on instruction-response data) language models, further enhance fluency and generalization in medical image-text tasks \cite{lee2025cxr,ye2024dynamic,zhang2024generalist}. But across CNN-RNN, Transformer-based, and LVLM-based methods, the predominant training paradigm remains token-level likelihood. This focus on local word prediction encourages stylistic mimicry, hallucinated but plausible statements, and incomplete coverage, and thus leaving report-level clinical correctness under-constrained and motivating objectives that act directly at the semantic, report level.

Building on this mismatch between token-level training and the needs of MRG, recent works added semantic supervision to better align text with image-based evidence (findings visually supported by the image). The most prominent method is via contrastive learning which aligns image-report pairs in a shared representation space. However, such approaches primarily provide global alignment signals at the report or image-text pair level, which are insufficient to capture local, fine-grained clinical phenomena, including specific findings and their entity-relation structure\cite{li2024contrastive,zhao2023medical,wang2021self}. Multi-task learning frameworks jointly train report generation with auxiliary classification or localization tasks, introducing explicit semantic supervision through predefined clinical labels. This strategy has been shown to improve coverage of labeled clinical findings and reduce unsupported statements by encouraging the model to mention image-grounded abnormalities. However, the effectiveness of multitask learning is constrained by the supervision itself: the available labels are often incomplete or noisy, the predefined label sets represent coarse clinical categories, and the additional prediction heads bias the model toward frequent, well-represented findings, making it less sensitive to rare but clinically important abnormalities \cite{jing2017automatic,wang2018tienet,wang2022automated}. A recent study on Dynamic Traceback Learning improves semantic alignment by explicitly linking generated report tokens back to their corresponding visual evidence, through masking or backtracking operations during training, which encourages the model to ground its text in the image. Nonetheless, the supervision is still provided through proxy training curricula, rather than direct evaluation of clinical correctness, and therefore does not explicitly optimize for clinical quality at the report level \cite{ye2024dynamic}. 

In this study, we propose a semantic-driven reinforcement learning (SRL) method for MRG that directly optimizes report-level clinical correctness, thereby improving the accuracy and coverage of clinically relevant findings, instead of optimizing token-level overlap that rewards local word matching. We instantiate this framework on a medical large vision-language model (Med-LVLM) to obtain MRG-R1, which is fine-tuned to generate reports that are better aligned with clinically meaningful findings, supported by explicit, self-generated reasoning. Rather than supervising individual tokens, we view the Med-LVLM as a stochastic report-generation model that samples complete reports given an input image, and optimize it using a report-level reward that reflects clinical semantics. To optimize non-differentiable, report-level clinical rewards efficiently for complete medical reports, we adopt a group-based policy optimization scheme, Group Relative Policy Optimization (GRPO)~\cite{shao2024deepseekmath}. This design specifically targets the gaps identified above: it shifts reward from local word-level patterns to clinically meaningful findings, encourages correct handling of positive and negative clinical statements grounded in the image, and introduces lightweight structural constraints to support interpretable reasoning.

We contribute to SRL by converting clinical semantics into a direct and stable report-level training signal, and coupling it with GRPO \cite{shao2024deepseekmath} to enable efficient optimization of complete medical reports without requiring an explicit value function. For each study, multiple candidate reports are sampled from the current policy; a CheXbert-based reward evaluates their clinical adequacy; and GRPO’s group-wise normalization amplifies the relatively best candidates for that case. In this way, MRG-R1 is explicitly encouraged to produce reports that are both clinically faithful and structurally organized, rather than merely stylistically similar to training texts.

Our main contributions can be summarized as follows:
\begin{itemize}
    \item \textbf{Semantic-Driven RL with GRPO for Clinically Aligned MRG.} 
We propose an SRL framework that optimizes MRG at the report level using clinically grounded rewards, addressing the mismatch between token-level training objectives and global clinical goals.
    \item \textbf{CheXbert-Guided Clinical Efficacy Reward and Instruction-Driven Explicit Reasoning.} 
We introduce MCCS, a margin-based CheXbert cosine-similarity reward, to score agreement in clinically meaningful findings, explicitly accounting for positive and negative observations while suppressing weak or incidental matches, thereby improving factual accuracy and reducing unsupported statements.
    \item \textbf{Comprehensive Empirical Experiments and Ablation Study.} Extensive experiments and analyses on the IU X-Ray \cite{demner2015preparing}, MIMIC-CXR \cite{johnson2019mimic}, have validate the  clinical efficacy of our method. 
\end{itemize}

\section{Related work}

\subsection{Medical Report Generation}

Medical report generation aims to produce sectioned narratives that capture anatomy, attributes, and clinically meaningful qualifiers within radiology reports. Beyond fluency, models must ensure factual adequacy, coverage of key findings at the report level.
Early MRG methods largely followed encoder-decoder designs from image captioning \cite{Ordonez:2011:im2text,vinyals2015show}, adapting them to longer, multi-sentence clinical reports. Jing et al. \cite{jing2017automatic} introduced hierarchical LSTMs with co-/cross-attention (plus auxiliary disease tags) to better plan sentences and ground wording in chest-X-ray findings. 
With Transformers, MRG shifted beyond recurrent decoders. R2Gen \cite{chen2020generating} introduced a memory-driven Transformer that caches global cues to model long-range context across report sections, achieving strong results on IU X-Ray and MIMIC-CXR, but still trained the generator with token-level objectives. Beyond R2Gen, R2GenCMN \cite{chen-acl-2021-r2gencmn} augments the memory-driven Transformer with cross-modal memory networks to strengthen image-text interactions under teacher forcing, yielding stronger token-level maximum likelihood estimation  (MLE) baselines on IU X-Ray and MIMIC-CXR. XProNet \cite{wang2022cross} further improves cross-modal interaction by learning shared visual-textual prototypes, enabling richer cross-modal pattern alignment for radiology report generation Subsequent variants injected prior and posterior knowledge (e.g., PPKED \cite{liu2021exploring}) to mitigate visual/textual biases while keeping the same learning target.  

Recently, Med-LVLMs have pushed capacity and generality. LLaVA-Med \cite{li2023llava} instruction-tunes a vision-language backbone on biomedical image-text pairs for dialogue, captioning, and VQA; Med-Flamingo \cite{moor2023med} adapts OpenFlamingo \cite{awadalla2023openflamingo} for few-shot generative medical VQA with physician blind review; HuatuoGPT-Vision \cite{chen2024huatuogpt} injects medical visual knowledge at scale into Qwen2-VL \cite{wang2024qwen2} and Qwen2.5VL \cite{bai2025qwen2} using an LLaVA-style training pipeline; MedGemma \cite{sellergren2025medgemma} adapts Gemma \cite{team2025gemma} to the medical domain via instruction tuning and evaluated biomedical tasks. CheXagent \cite{chen2024chexagent} targets chest X-ray interpretation and multi-task evaluation via a curated instruction datasets. Radiology-focused variants such as CXR-LLaVA \cite{lee2025cxr} tailor LLaVA \cite{liu2023visual} to chest X-rays and study zero-/few-shot reporting or recognition; broader “generalist” biomedical models (e.g., BioMedGPT \cite{zhang2024generalist}) and radiology foundation efforts (e.g., RadFM \cite{wu2025towards}) pursue unified pretraining across modalities. 

In summary, medical report generation has progressed from early encoder-decoder models to Transformer-based architectures and more recently instruction-tuned LVLMs. These advances substantially improve fluency and overall report structure, yet most systems are still optimized primarily with token-level maximum likelihood objectives. As a result, training tends to favor local lexical matching rather than directly enforcing clinically grounded, report-level semantics. 

\subsection{Semantic Supervision} 


To mitigate the mismatch between token-level objectives and clinical goals, a growing amount of work augments report generation with semantic supervision that more directly reflects clinical meaning and image grounding. 
Knowledge-based approaches incorporate external knowledge to guide generation. For example, GSKET \cite{yang2022knowledge} integrates general (graph-based) and case-specific knowledge to support sentence planning, and Dynamic Graph Enhanced Contrastive Learning (DCL) \cite{li2023dynamic} refines graph structure and introduces contrastive objectives to encourage finer-grained semantic alignment.

A complementary thread strengthens image-text grounding via matching or contrastive signals. Co-training a generator with image-text matching (ITM) heads (“self-boosting” \cite{wang2021self}) improves clinical alignment by penalizing mismatched pairs. Reinforced Cross-modal Alignment \cite{qin2022reinforced} introduces an RL objective over a cross-modal memory to better couple visual and textual cues; and segment-enhanced contrastive learning (MSCL \cite{zhao2023medical}) leverages segmentation to focus alignment on clinically meaningful regions of interest and reduce dataset bias. Retrieval-assisted systems operationalize this alignment at inference time: X-REM \cite{jeong2024multimodal} learns a contrastive matching score to retrieve report sentences conditioned on the image, improving grounding and reducing unsupported statements. CXRMate \cite{nicolson2024longitudinal} introduces longitudinal semantic rewards that leverage follow-up consistency signals to reduce hallucinations and improve clinically coherent reporting in chest X-rays.

Moving closer to clinically grounded supervision, labeler-based methods turn automatic clinical labelers into training or evaluation signals. CheXbert \cite{smit2020chexbert}, a BERT-based radiology report labeler, extracts 14 chest observations and their presence/absence status, and is widely used to score clinical efficacy and provide label-level guidance for report generation.

However, existing semantic supervision is often indirect: contrastive and matching losses provide global alignment, multitask classifiers depend on coarse or noisy labels, and traceback-style methods rely on proxy curricula rather than explicit clinical rewards. This leaves an open need for polarity-sensitive, report-level objectives that can be directly optimized during generation.

\subsection{Reinforcement Learning}



Post-training for LLMs/LVLMs increasingly relies on preference-based objectives \cite{ouyang2022training,bai2022constitutional}. The standard Reinforcement Learning from Human Feedback (RLHF) \cite{ouyang2022training} framework trains a reward model from human preference comparisons and then optimizes the policy using Proximal Policy Optimization (PPO) \cite{schulman2017proximal} with a Kullback-Leibler (KL) divergence constraint \cite{kullback1951kullback}. This recipe was popularized by InstructGPT, and PPO's clipped surrogate objective provides stable on-policy updates. To cut human labeling, RLAIF/Constitutional AI \cite{bai2022constitutional} uses AI feedback and rule-based critiques. Direct Preference Optimization (DPO) \cite{rafailov2023direct} simplifies RLHF by removing the explicit reward-model and value-function training loop and directly optimizing on preference pairs. Group Relative Policy Optimization (GRPO) \cite{shao2024deepseekmath} estimates group-relative advantages from multiple sampled responses and performs updates without learning an explicit value function, which makes it more memory-efficient and effective for improving model reasoning. Together, these methods reflect a broader shift from supervised next-token fitting toward preference-driven post-training.

Prior work thus demonstrates the value of sequence-level and preference-based optimization, but has rarely targeted clinically grounded rewards for long-form radiology reports. We build on GRPO in this setting, coupling it with a rule/labeler-based clinical reward to align Med-LVLMs with report-level medical correctness.

\section{Method}

\begin{figure*}[!t]  
  \centering
  \resizebox{0.85\linewidth}{!}{
  \includegraphics[width=\textwidth]{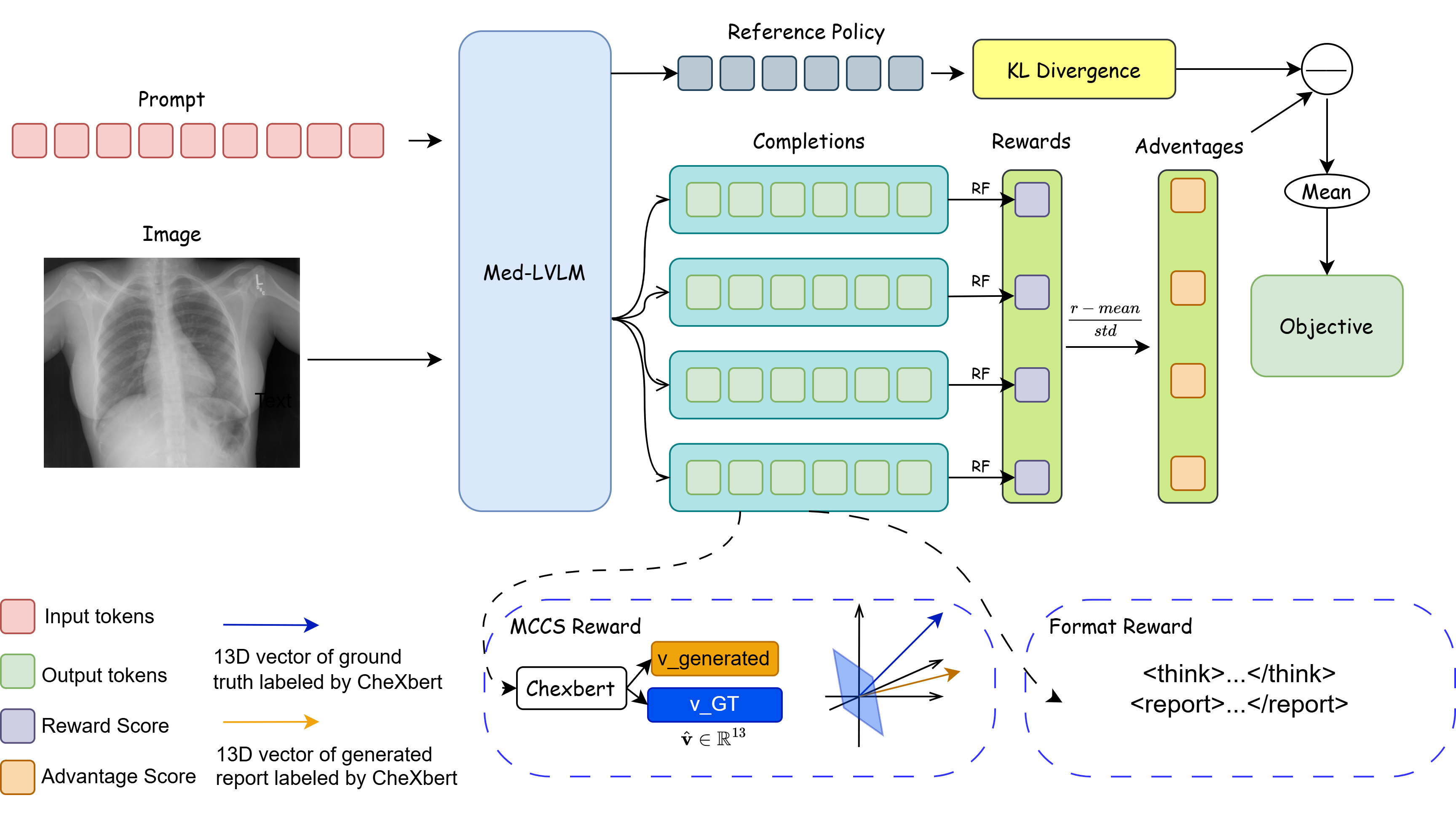} }
  \caption{Overview of SRL. For each study, the policy samples a group of candidate reports; a margin CheXbert cosine reward (MCCS) and a lightweight format reward are combined to compute group-relative advantages for GRPO updates under a KL constraint to a reference policy.}
  \label{fig:method_overview}
\end{figure*}

We cast post-training for MRG as reinforcement learning over a clinically grounded, report-level reward. As shown in Figure \ref{fig:method_overview} given a chest X-ray, the Med-LVLM (policy) samples multiple candidate reports; CheXbert-derived label vectors define a margin-based cosine reward (MCCS), and a lightweight format reward checks the \verb|<think>→<report>| structure. These rewards are combined within groups and optimized via GRPO under a KL constraint to a reference policy. We fine-tune the LVLM with GRPO \cite{shao2024deepseekmath}, a value-free, group-wise policy-gradient method selected for its stability and compute efficiency when optimizing sequence-level, report-level clinical rewards.  The optimization signal is a CheXbert-guided reward that scores agreement across 14 chest-x-ray observations between the generated report and the reference, providing direct supervision on clinical content and reducing the reliance on n-gram overlap. We next detail (i) the GRPO training loop: sampling, group-relative advantage computation, and update rule. (ii) the reward functions, including the CheXbert margin-cosine design and its aggregation at the report level. 

\subsection{Group Relative Policy Optimization (GRPO)}

GRPO \cite{shao2024deepseekmath} is a PPO-style \cite{schulman2017proximal} post-training algorithm that optimizes reward-defined objectives instead of pure likelihood. In our setting, the LVLM is treated as a policy that generates full reports, and GRPO updates this policy using group-relative advantages computed from our clinical rewards. We leverage this to bias generation toward clinically aligned report-level targets, providing direct supervision on semantic fidelity beyond token overlap. GRPO is closely related to PPO but differs in two key aspects: first, GRPO estimates the advantage using group-based estimation rather than a value function; second, it uses fixed task-specific reward functions (CheXbert-based and format rewards) instead of a learned value network.

Let $P(Q)$ denote the training set of inputs (“studies”); a single input is $q \in P(Q)$ , We write $\pi_{\theta_{\text{old}}}$ and $\pi_{\theta_{\text{new}}}$ for the old policy (used to sample responses in the current update) and the current policy (parameters being optimized), respectively. A complete response $o$ means the full generated report for $q$. We also use a frozen reference policy $\pi_{\theta_{\text{ref}}}$, to regularize updates. Let $G$ be the group size, the number of responses sampled per input $q$ at each iteration, yielding $\{o_i\}_{i=1}^{G}$. 

The GRPO objective is 

\begin{equation}
\begin{aligned}
J_{\mathrm{GRPO}}(\theta)
&= \mathbb{E}_{\,q\sim P(Q),\;\{o_i\}_{i=1}^{G}\sim \pi_{\theta_{\mathrm{old}}}(\cdot \mid q)} \\
&\quad \Bigg[
\frac{1}{G}\sum_{i=1}^{G}
\min\!\Bigg(
\frac{\pi_{\theta_{\mathrm{new}}}(o_i \mid q)}{\pi_{\theta_{\mathrm{old}}}(o_i \mid q)}\,A_i,\;\\
&\quad
\operatorname{clip}\!\Big(
\frac{\pi_{\theta_{\mathrm{new}}}(o_i \mid q)}{\pi_{\theta_{\mathrm{old}}}(o_i \mid q)},
\,1-\epsilon,\,1+\epsilon
\Big)\,A_i
\Bigg)
 \\
&\quad - \beta\,D_{\mathrm{KL}}\!\big(\pi_{\theta_{\mathrm{new}}}\,\big\|\,\pi_{\theta_{\mathrm{ref}}}\big) \Bigg]
\end{aligned}
\label{eq:grpo_objective}
\end{equation}
Here, the policy ratio $\frac{\pi_{\theta_{\text{new}}}(o_i\mid q)}{\pi_{\theta_{\text{old}}}(o_i\mid q)}$ measures how the new policy probability of $o_i$ changes relative to the old policy; $A_i$ is the estimated advantage for response $o_i$; $\epsilon > 0$ is the clipping threshold that limits overly large updates by replacing the raw ratio with its clipped version; and $D_{\mathrm{KL}}(P\|Q)$ is the  KL divergence \cite{kullback1951kullback} between the new policy and the reference policy, scaled by $\beta \geq 0$ to control policy drift. Intuitively, the “$min$” enforces the clipped surrogate familiar from PPO, while the KL term keeps the updated policy close to $\pi_{\theta_{\text{ref}}}$.

Unlike PPO, which estimates $A_i$ using a learned value function (critic), GRPO computes $A_i$ from the sampled responses within the same group for a given input, thereby avoiding explicit value estimation. Concretely, with rewards $r_i = R(q,o_i)$ from our rule/labeler-based clinical reward $R$, we use a normalized, within-group advantage 
\begin{equation}
\begin{aligned}
\bar r &= \frac{1}{G}\sum_{i=1}^{G} r_i,\\
\sigma_r &= \sqrt{\frac{1}{G}\sum_{i=1}^{G}(r_i-\bar r)^2+\varepsilon},\\
A_i &= \frac{r_i-\bar r}{\sigma_r}.
\end{aligned}
\label{eq:grpo_adv}
\end{equation}
where $\bar r$ and $\sigma_r$ are the group mean and standard deviation, Here, $\varepsilon>0$ is a small constant for numerical stability; when the group rewards are identical (variance near zero), set $\varepsilon$ to a value on the order of $10^{-8}$–$10^{-6}$. This relative construction compares candidates conditioned on the same study $q$. sharpening the learning signal for report-level clinical rewards without training a critic.

\subsection{Reward Functions}

\subsubsection{Format Reward}

We use a format reward to elicit explicit, auditable reasoning without requiring CoT annotations. The prompt asks the model to place intermediate reasoning inside \verb|<think>…</think>| and the final radiology report inside \verb|<report>…</report>|. A rule-based scorer evaluates only structure: tags must be present, correctly ordered, well-formed (balanced), and non-empty. Outputs that fully comply receive a score of 1, with partial credit for minor violations; otherwise the score is 0. This term is added with a small weight relative to the clinical reward so optimization remains driven by medical correctness. Under GRPO’s group-relative updates, candidates that satisfy the structure reliably obtain higher relative advantages within the same case, teaching the policy to produce a stable two-stage “reasoning → report” format. The benefits are threefold: (i) decoupling thinking from the final narrative, (ii) improving readability and downstream parsing, and (iii) enabling auditability to localize hallucinations or inconsistencies.

\subsubsection{Margin Chexbert Cosine Similarity Reward (MCCS)}

Beyond enforcing output structure via the format reward, our optimization is driven primarily by a clinically grounded signal that evaluates report-level semantics. We instantiate this signal as a Margin CheXbert Cosine Similarity (MCCS) reward, which converts CheXbert’s 14-label \cite{smit2020chexbert} outputs into signed vectors and rewards their margin-calibrated cosine agreement, providing a continuous target for GRPO. Let CheXbert provide a 14-way multi-class label over common chest-X-ray observations for each study. We map each observation to a scalar by
\begin{equation}
\label{eq:mapping}
f(\text{pos})=1,
f(\text{neg})=-1,
f(\text{uncertain})=1,
f(\text{blank})=0.
\end{equation}

and construct report-level vectors $\mathbf{z}(y), \mathbf{z}(y^\star) \in \mathbb{R}^{13}$ over the 13 disease-specific categories only (exclude No Finding) for the generated report \(y\) and the reference $y^\star$. 
\begin{equation}
z_j(y)=f\big(\ell_j(y)\big),
z_j(y^\star)=f\big(\ell_j(y^\star)\big),\quad j=1,\dots,13 .
\label{eq:mccs_vec}
\end{equation}
Mapping uncertain to 1 treats hedged mentions as actionable suspicion rather than neutrality, which matches clinical practice: when radiologists hedge, they are flagging a possible abnormality that warrants attention. By contrast, blank is 0, reflecting true omission. This choice biases the reward toward sensitivityit favors correctly surfacing potential findings and still penalizes polarity reversals (positive vs. negative) most strongly via the signed embedding. It also discourages “safe” under-calling: labeling everything as uncertain no longer evades penalties if the reference is negative (–1) or omitted (0), and it earns credit only when uncertainty aligns with a true or suspected abnormality. We also exclude the No Finding dimension from the cosine similarity. In CheXbert \cite{smit2020chexbert}, No Finding is typically set to 1 when all other disease labels are 0. As a result, it can dominate vector norms and inflate apparent agreement via complementarity, and it is highly sensitive to reporting style or templated omissions, thereby introducing noise. Removing this dimension focuses the signal on per-finding clinical agreement and avoids pseudo-alignment driven by a global catch-all label.

We then measure report-level agreement via cosine similarity 
\begin{equation}
\mathrm{CCS}(y,y^\star)=
\frac{\langle z(y),z(y^\star)\rangle}{\big(\lVert z(y)\rVert_{2}+\varepsilon\big)\big(\lVert z(y^\star)\rVert_{2}+\varepsilon\big)},\quad \varepsilon=10^{-8}.
\label{eq:cosine}
\end{equation}
This $\varepsilon$ guarantees numerical safety even when one vector is (nearly) zero after our preprocessing (e.g., with the No Finding dimension removed), while leaving values effectively unchanged when norms are in a normal range. To calibrate the signal and emphasize clinically meaningful improvements, we convert cosine similarity to a margin-shaped reward: 
\begin{equation}
\mathrm{MCCS}(y,y^\star,m)
=\max\left(\frac{\mathrm{CCS}(y,y^\star)-m}{1-m},0\right), m\in(-1,1).
\label{eq:mccs}
\end{equation}
This piecewise-linear shaping has three advantages. (i) Margin filtering. Scores at or below $m$ yield zero reward, suppressing weak alignments (e.g., incidental overlap) and focusing learning on clinically aligned matches. (ii) Dynamic-range normalization. The division by $(1-m)$ maps $CCS \in [m,1]$ to $[0,1]$, ensuring comparable reward scales across studies and increasing within-group variance when $m$ is moderate—beneficial for GRPO’s group-relative advantages. (iii) Stable gradients. The linear slope $1/(1-m)$ avoids early saturation near high similarity and provides smooth, interpretable shaping; $MCCS = 1$ if and only if the two label vectors coincide up to positive scaling. 

In all cases, MCCS acts as a continuous, clinically grounded reward at the report level, providing partial credit for near matches and stronger penalties for polarity mistakes than for uncertainty/omission, thereby aligning optimization with clinical correctness rather than token overlap.

\section{Experimental Setup}

\subsection{Datasets}

MIMIC-CXR \cite{johnson2019mimic} contains 473,057 chest X-ray images and 227,835 radiology reports. For comparison with prior works, we adopt the split provided by MIMIC-CXR with approximately 222.8k/1.8k/3.3k samples for training/validation/test following \cite{yang2022knowledge,chen2022cross}. IU X-Ray \cite{demner2015preparing} comprises 7,470 images and 3,955 reports. We follow \cite{chen2020generating,chen2022cross} and use a 70/10/20 train/validation/test split. Unless otherwise noted, multi-view studies (reports associated with multiple images) are treated as multiple image-report pairs, with each image paired to the same report and counted as a separate sample. 

\subsection{Implementation Details}

All experiments are conducted on 2×NVIDIA A100 GPUs. We fine-tune HuatuoGPT-Vision-7B-Qwen2.5VL
 \cite{chen2024huatuogpt}, a Qwen2.5-VL-based \cite{bai2025qwen2} vision-language model further trained on medical image-text and instruction data. We adopt parameter-efficient LoRA tuning \cite{hu2022lora} (rank 128, $\alpha=256$, dropout 0.05) on attention and MLP projections, with FlashAttention-2 \cite{dao2022flashattention} and bfloat16 for memory efficiency. Optimization uses 8-bit AdamW \cite{loshchilov2017decoupled} with learning rate $5\times10^{-6}$, $(\beta_1,\beta_2)=(0.9,0.99)$, weight decay 0.1, cosine decay with 10\% warm-up, gradient clipping at 0.1, effective batch size 16, and 1 training epoch. We apply DeepSpeed ZeRO-1 \cite{rasley2020deepspeed} for optimizer sharding. For GRPO, each input samples $G=4$ candidate reports; group-relative advantages are computed within the group, and, when combining rewards, the total reward is a weighted sum of clinical (0.75) and format (0.25) terms.

\subsection{Evaluation Metrics}

We evaluated the quality of generated reports using clinical efficacy (CE) metrics that reflect factual correctness rather than stylistic similarity. Specifically, we adopt CheXbert-based precision, recall, and F1 computed over 14 chest X-ray observations defined by CheXbert, following the standard evaluation protocol in prior work\cite{ye2024dynamic,yang2022knowledge,chen2024chexagent,smit2020chexbert}. In contrast, conventional NLG metrics such as BLEU, ROUGE, and CIDEr primarily reward n-gram overlap and template reuse, which often obscure factual adequacy and fail to penalize polarity errors. Multiple studies have shown that such lexical metrics correlate weakly with radiologists’ judgments of factual accuracy, whereas CE metrics better track clinically relevant errors\cite{yu2023evaluating,ostmeier2024green,liu2024mrscore}.

\subsection{Baselines}

In this study, we compare MRG-R1 with three families of baselines in IU X-Ray and MIMIC-CXR, using code and checkpoints released when available and retraining with the authors' settings otherwise. (A) \emph{Token-level MLE generators}: R2Gen \cite{chen2020generating}, R2GenCMN \cite{chen-acl-2021-r2gencmn} and XProNet \cite{wang2022cross} representative encoder-decoder and transformer models trained under teacher forcing. (B) \emph{Instruction-tuned Med-LVLMs}: BioMedGPT \cite{zhang2024generalist}, LLaVA-Med \cite{li2023llava}, CheXagent \cite{chen2024chexagent} and HuatuoGPT-Vision \cite{chen2024huatuogpt}, evaluated under a uniform prompting and decoding setup without additional fine-tuning on our splits to probe zero-/few-shot reporting ability and domain alignment. (C) \emph{Semantic supervision}: DTrace \cite{ye2024dynamic}, DCL \cite{li2023dynamic}, GSKET \cite{yang2022knowledge}, CXRMate \cite{nicolson2024longitudinal}, and RadFM \cite{wu2025towards}, which inject clinical semantics through traceback, contrastive, knowledge graphs or radiology-focused pre-training. Note that some baselines (e.g., CheXagent) are instruction-tuned on substantially broader medical corpora beyond MIMIC-CXR and IU X-Ray, and their performance may partly reflect pretraining coverage rather than architecture alone, therefore we treat them as strong external baselines rather than strictly comparable models.

\section{Results}

\subsection{Quantitative Analysis}

We compared MRG-R1 against existing report generation models on IU-Xray and MIMIC-CXR. 

\begin{table*}[t]
\centering
\resizebox{0.85\linewidth}{!}{
\begin{tabular}{l|c|ccc|ccc}
\hline
\multicolumn{1}{c|}{\multirow{2}{*}{Method}} & \multirow{2}{*}{\begin{tabular}[c]{@{}c@{}}Traning \\ Technology\end{tabular}}           & \multicolumn{3}{c|}{IU X-Ray}                    & \multicolumn{3}{c}{MIMIC-CXR}                    \\ \cline{3-8} 
\multicolumn{1}{c|}{}                        &                                                                                          & Precision      & Recall         & F1             & Precision      & Recall         & F1             \\ \hline
R2Gen \cite{chen2020generating}              & \multirow{3}{*}{\begin{tabular}[c]{@{}c@{}}Token-level\\ MLE \\ generators\end{tabular}} & \uline{50.60}* & 48.76*         & 46.99*         & 33.30          & 27.30          & 27.60          \\
R2GenCMN \cite{chen-acl-2021-r2gencmn}       &                                                                                          & 50.00*         & 51.07*         & 50.53*         & 33.40          & 27.50          & 27.80          \\
XProNet \cite{wang2022cross}                 &                                                                                          & --             & --             & --             & 30.20          & 22.20          & 25.50          \\ \hline
RadFM \cite{wu2025towards}                   & \multirow{5}{*}{\begin{tabular}[c]{@{}c@{}}Instruction-tuned\\ Med-LVLMs\end{tabular}}   & 14.27*         & 11.93*         & 12.99*         & 10.03*         & 12.08*         & 10.96*         \\
BioMedGPT \cite{zhang2024generalist}         &                                                                                          & 36.00*         & 35.40*         & 35.50*         & 29.00*         & 31.40*         & 28.60*         \\
LLaVA-Med \cite{li2023llava}                 &                                                                                          & 18.63*         & 23.37*         & 20.73*         & 26.99*         & 12.03*         & 16.64*         \\
CheXagent \cite{chen2024chexagent}           &                                                                                          & 50.37*         & \uline{51.96}* & \uline{51.15}* & 45.60*         & 24.59*         & 31.95*         \\
HuatuoGPT-Vision \cite{chen2024huatuogpt}    &                                                                                          & 5.87*          & 7.33*          & 6.52*          & 23.67*         & 16.51*         & 19.45*         \\ \hline
DTrace \cite{ye2024dynamic}                  & \multirow{5}{*}{\begin{tabular}[c]{@{}c@{}}Semantic \\ supervision\end{tabular}}         & --             & --             & --             & 41.10          & \textbf{43.60} & \uline{39.10}  \\
DCL \cite{li2023dynamic}                     &                                                                                          & --             & --             & --             & \textbf{47.10} & 35.20          & 37.30          \\
GSKET \cite{yang2022knowledge}               &                                                                                          & --             & --             & --             & \uline{45.80}  & 34.80          & 37.10          \\
CXRMate \cite{nicolson2024longitudinal}      &                                                                                          & 28.30          & 35.10          & 27.70          & 43.80          & 34.90          & 35.70          \\
MRG-R1 (ours)                                &                                                                                          & \textbf{50.86} & \textbf{52.98} & \textbf{51.88} & 45.32          & \uline{37.70}  & \textbf{40.39} \\ \hline
\end{tabular}}
\caption{Clinical efficacy (CE) comparison on IU X-Ray and MIMIC-CXR. 
    CE is computed with CheXbert over the 14 standard observations. 
    *  denote scores is not provided by author and reproduced by our. \textbf{Bold} marks the best performance per column; \uline{underline} marks the second best. 
    For IU X-Ray, CE is obtained by running CheXbert on both the generated and the ground-truth reports, consistent with prior work. Unless otherwise stated, CE is computed on all 14 CheXbert observations (including \emph{No Finding}); MCCS used for SRL excludes \emph{No Finding} as described in Section ~3.}
    \label{tab:ce_main}
\end{table*}

Across both datasets, our MRG-R1 delivered strong clinical efficacy (CE), achieving state-of-the-art performance on IU X-Ray by attaining the highest F1 (51.88), edging out classical encoder-decoder baselines such as R2GenCMN (50.53) and exceeding the top Med-LVLM systems (e.g., CheXagent 51.15). The gains were driven by balanced improvements in both precision (50.86) and recall (52.98), suggesting that SRL with GRPO enhanced sensitivity to clinically salient findings such as key abnormalities and attributes that are essential for radiological assessment and decision-making, while still maintaining a low false-positive rate. 

On MIMIC-CXR, MRG-R1 achieved the best F1 in Table, reaching 40.39 and substantially outperforming classical MLE baselines such as R2GenCMN (27.80). It also outperformed prior semantic-supervision methods, including DTrace, DCL, GSKET, and CXRMate, as well as instruction-tuned or generalist medical LVLMs such as LLaVA-Med, BioMedGPT, and HuatuoGPT-Vision. These results further demonstrate the value of optimizing a clinically grounded report-level reward instead of relying solely on token-level imitation or generic instruction tuning.

We further observed that methods injecting semantic signals without RL (e.g., DTrace’s traceback supervision; CXRMate’s longitudinal semantic reward) yielded stronger CE than early MLE systems, but MRG-R1 remained competitive or superior on average while using a compute-efficient GRPO update without an explicit value function. 

\subsection{Qualitative Analysis}
\begin{figure*}[]  
    \centering
    \includegraphics[width=0.9\linewidth]{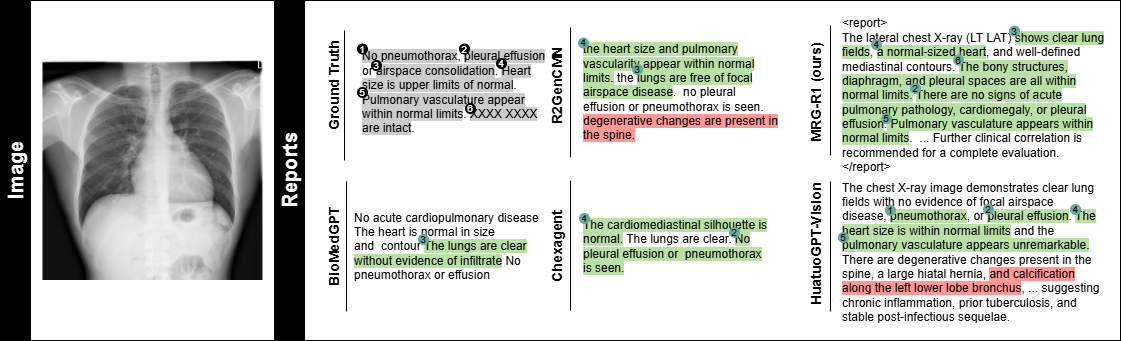}
    \caption{An example case from IU X-ray (X-ray image) used for inference in MRG qualitative comparisons. The information
 in the ground truth report is labeled from 1 to 6 and highlighted separately. The generated reports are labeled according to the ground truth report and
 high lighted with different colors to represent the differences between the generated sequences and the ground truth report: (1)Green-consistent; (2)Red- incorrect information; (3)Unhighlighted-not included in the ground truth. }
    \label{fig:iu_casestudy}
\end{figure*}

\begin{figure*}[]  
    \centering
    \includegraphics[width=0.9\linewidth]{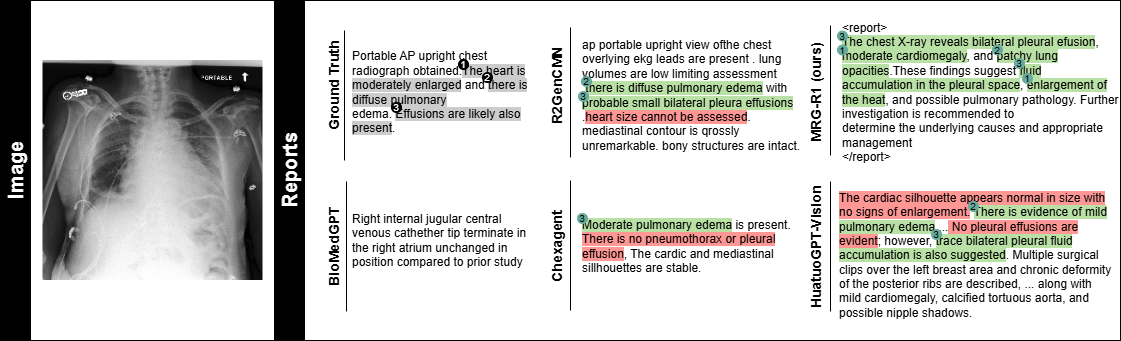}
    \caption{An example case from MIMIC-CXR (X-ray image) used for inference in MRG qualitative comparisons. The information
 in the ground truth report is labeled from 1 to 3 and highlighted separately. The generated reports are labeled according to the ground truth report and
 high lighted with different colors to represent the differences between the generated sequences and the ground truth report: (1)Green-consistent; (2)Red- incorrect information; (3)Unhighlighted-not included in the ground truth. }
    \label{fig:mimic_casestudy}
\end{figure*}

We compared report outputs on representative IU X-Ray and MIMIC-CXR studies (Figures \ref{fig:iu_casestudy} - \ref{fig:mimic_casestudy}). Both examples illustrate four dimensions that drive clinical utility: (i) polarity handling meaning whether the report states the presence or absence of important abnormalities accurately, (ii) handling of uncertainty where equivocal findings are expressed as suspicion rather than definite statements, (iii) the balance between omissions and hallucinations, avoiding both missing critical findings and introducing unsupported ones , and (iv) structural coherence where the report is organized in a clear radiology-style format that separates reasoning from conclusions.

On IU X-Ray (Fig.\ref{fig:iu_casestudy}), the ground-truth report emphasizes normal lungs and pleura with a heart size at the upper limit of normal. MRG-R1 aligns most closely with this reference by preserving the correct polarity of the major findings, including clear lungs, no pleural effusion or pneumothorax, and no cardiomegaly. By contrast, the baselines show different error types: BioMedGPT is relatively fluent but generic and incomplete, R2GenCMN adds non-central skeletal findings, CheXagent under-describes the case, and HuatuoGPT-Vision introduces several unsupported abnormalities. 

On MIMIC-CXR (Fig.\ref{fig:mimic_casestudy}), the ground truth report documents cardiomegaly, pulmonary edema, and likely effusions as the abnormality. MRG-R1 recovers all three abnormalities with the correct polarity. By contrast, the baselines exhibit different failure modes: BioMedGPT emphasizes device position while omitting the principal pathology; R2GenCMN gives a less specific description of cardiac enlargement and pleural abnormality; CheXagent captures edema but incorrectly negates pleural effusion; and HuatuoGPT-Vision misses the pleural fluid finding and also states that the cardiac silhouette is normal in size. Overall, MRG-R1 aligns most closely with the reference report in this case. 

\subsection{Ablation Studies}

We conducted ablation studies on our MRG-R1 to assess the contribution of each component in our method. Table~\ref{tab:ablation} presents the ablated results comprising: (1) supervised fine-tuning (cross-entropy), (2) text-level NLG rewards (BLEU+ROUGE+CIDEr), (3) a format-only reward that enforces a \texttt{<think>}$\rightarrow$\texttt{<report>} structure (Format), (4) a clinical reward via report-level CE-F1 (with/without Format), and (5) our margin CheXbert cosine similarity (MCCS, with/without Format). This sequence disentangles stylistic supervision, structural guidance, and clinically grounded objectives. 

\begin{table*}[t]
  \centering
  \renewcommand{\arraystretch}{1.12}
  \setlength{\tabcolsep}{6pt}
  \begin{tabular}{l|ccc|ccc}
\hline
\multirow{2}{*}{Method}             & \multicolumn{3}{c|}{IU X-Ray}                    & \multicolumn{3}{c}{MIMIC-CXR}                    \\ \cline{2-7} 
                                    & Precision      & Recall         & F1             & Precision      & Recall         & F1             \\ \hline
\makecell[l]{Base}                  & 5.87           & 7.33           & 6.52           & 23.67          & 16.51          & 19.45          \\
\makecell[l]{Base + SFT}            & 3.86           & 7.09           & 4.99           & 24.27          & 15.00          & 14.64          \\
\makecell[l]{Base + NLG}            & 41.19          & 15.93          & 22.97          & 24.74          & 8.56           & 12.72          \\
\makecell[l]{Base + CE-F1}          & 45.89          & 43.78          & 44.81          & 36.38          & 25.08          & 29.69          \\
\makecell[l]{Base + format}         & 24.21          & 38.33          & 29.67          & 27.71          & 25.53          & 26.58          \\
\makecell[l]{Base + CE-F1 + format} & 50.04          & \uline{52.73}    & \uline{51.35}    & 33.00          & 28.87          & 29.50          \\
\makecell[l]{Base + MCCS}           & \textbf{53.27} & 46.51          & 49.66          & \uline{36.07}    & \textbf{44.69} & \uline{38.67}    \\
\makecell[l]{Base + MCCS + format}  & \uline{50.86}    & \textbf{52.98} & \textbf{51.88} & \textbf{45.32} & \uline{37.70}    & \textbf{40.39} \\ \hline
\end{tabular}
  \caption{Ablation on IU X-Ray and MIMIC-CXR starting from a zero-shot HuatuoGPT-Vision-7B (Base). 
We incrementally add: supervised fine-tuning (cross-entropy), text-level NLG rewards (BLEU+ROUGE+CIDEr), a format-only reward enforcing a \texttt{<think>} $\rightarrow$ \texttt{<report>} structure (Format), a clinical reward via report-level CE-F1 (with/without Format), and our margin CheXbert cosine similarity (MCCS, with/without Format). 
\textbf{Bold} marks the best per column; \uline{underline} the second best.}

  \label{tab:ablation}
\end{table*}

SFT alone does not consistently improve clinical efficacy (CE) (IU: F1 6.52$\rightarrow$4.99; MIMIC: 19.45$\rightarrow$14.64), likely because cross-entropy training remains weakly aligned with CheXbert-based clinical labels and can overfit to surface phrasing without improving label-level correctness. Relative to Base, optimizing purely lexical NLG rewards (\,+NLG\,) improves fluency but yields limited clinical efficacy (CE): F1 rises only to 22.97 on IU X-Ray and 12.72 on MIMIC-CXR, consistent with the weak linkage between n-gram overlap and factual correctness. Replacing the objective with a clinical signal (\,+CE-F1\,) substantially improves CE (IU 44.81; MIMIC 29.69), indicating that label-consistency supervision reduces polarity errors and under-calling. A format-only constraint (\,+Format\,) increases recall (IU 38.33; MIMIC 25.53) at some cost to precision, while \,+CE-F1+Format\, stabilizes negation/uncertainty templates and recovers a strong precision-recall balance (IU F1 51.35).

Our MCCS is the most effective shaping in this setting. Compared with CE-F1, MCCS maps CheXbert labels to signed vectors (pos\,=\,1, neg\,=\,-1, blank\,=\,0, uncertain\,=\,1), excludes the catch-all \emph{No Finding}, and applies a margin that suppresses weak matches. This polarity-sensitive, sequence-level signal widens case-level score separation, which GRPO’s group-relative updates leverage to amplify the best candidate per study. Empirically, \,+MCCS\, boosts recall on MIMIC-CXR (44.69; second-best F1 38.67), and \,+MCCS +Format\, delivers the best overall CE on both datasets (IU F1 51.88; MIMIC F1 40.39) with the highest MIMIC precision (45.32). These trends support MCCS as a stronger clinical reward than CE-F1 under GRPO and motivate pairing it with a light format constraint for stable long-form generation.

\section{Discussion}

Our experiments show that MRG-R1 achieved state-of-the-art on IU-Xray and strong performance on MIMIC-CXR under our unified evaluation protocol. These results indicate that optimizing report-level, clinically grounded rewards via reinforcement learning can improve label-level clinical correctness compared with token-level maximum likelihood training. Specifically, by shifting the training objective from next-token likelihood to a report-level clinical reward, the model is explicitly encouraged to produce reports with more accurate affirmed vs. negated findings and better coverage of key observations. As reflected in the improved precision-recall balance, correct affirmations/negations are rewarded while omissions and unsupported statements are discouraged. In contrast, conventional maximum-likelihood training mainly reinforces locally plausible word sequences, which can improve fluency but does not directly optimize report-level clinical adequacy.

The improvement observed on IU X-Ray suggests that GRPO-based reinforcement learning with report-level rewards can be effective when supervision is limited. Compared with MIMIC-CXR (227,835 studies/377,110 images), IU X-Ray is much smaller dataset (3,955 studies/7,470 images) and and its reports are shorter and less diverse, which may make reward-based optimization more sample-efficient \cite{wu2023exploring}. 
In this setting, optimizing a report-level clinical reward provides a stronger training signal than token-level likelihood: for each study, the model generates multiple candidate reports, and updates are driven by which candidates better match clinically relevant findings under the reward. This encourages the model to focus on improving clinically meaningful content (e.g., affirmed vs. negated findings and coverage of key observations), rather than only matching local wording patterns. On the larger and more heterogeneous MIMIC-CXR benchmark, the absolute gains are smaller but remain substantial, indicating that the same reward-based optimization continues to improve clinical efficacy under a more complex data distribution.

We further analyze the contribution of each component through the ablation study in Table~\ref{tab:ablation}. Starting from the zero-shot HuatuoGPT-Vision base model, conventional supervised fine-tuning with cross-entropy (+SFT) does not improve and can even reduce CE, illustrating that on these datasets MLE primarily reinforces stylistic patterns rather than clinically grounded semantics. When we switch to GRPO but optimize only NLG metrics (+NLG), the model learns to adjust style and surface overlap, but CE remains low especially on MIMIC-CXR which confirming that BLEU/ROUGE/CIDEr are poor optimization targets for clinical correctness in MRG. The format-only reward (+Format), which enforces the “\verb|<think>…</think>→<report>…</report>|” structure without any clinical reward, increases recall but reduces precision, consistent with the model mentioning more findings without being penalized for unsupported positives. Introducing a CheXbert-based CE-F1 reward (+CE-F1) substantially improves CE, and combining it with the format constraint (+CE-F1 +Format) yields a better precision-recall balance, indicating that label-level supervision helps reduce omissions and improve agreement in affirmed vs. negated findings. 
Compared with CE-F1, our MCCS reward provides a smoother report-level signal by scoring CheXbert label agreement, excluding the noisy “No Finding” dimension, and suppressing weak matches via a margin. This shaping increases separation among candidate reports for the same study, which GRPO can exploit via group-relative updates. When combined with the format constraint (+MCCS +Format), the model achieves the strongest and most balanced CE on both datasets while producing more consistently structured outputs. 

Qualitative comparisons further illustrate how SRL reshapes model behaviour beyond CE scores alone. In the IU X-Ray and MIMIC-CXR case studies, MRG-R1 more consistently handles affirmed versus negated findings. This suggests that the reward discourages label-level inconsistencies (e.g., switching a finding between present and absent) rather than only improving fluency. The model also tends to surface clinically important findings instead of defaulting to generic “no acute abnormality” templates, suggesting that sequence-level optimization encourages it to prioritise label-supported observations over safe but uninformative normal statements. Its handling of uncertainty is more calibrated: equivocal patterns are framed as suspicion rather than as definite disease or definite normality, which aligns with the way radiologists hedge when evidence is borderline and reflects the polarity- and uncertainty-aware design of MCCS. The \verb|<think>→<report>| structure additionally makes the reasoning process more transparent; intermediate “thinking” appears to mirrors the CheXbert label space, while the final report rephrases those decisions into radiology-style prose, providing a natural point for auditing where hallucinations or misinterpretations arise. At the same time, the residual errors we observe: missed subtle or highly localized findings, incomplete description of devices and lines, and overly conservative summaries in ambiguous cases are concentrated precisely in areas where the current reward is blind or weak, namely phenomena not covered by the 14-label scheme or poorly captured by single-time-point views. This pattern supports the view that MRG-R1 is genuinely aligning to the supervision it receives, rather than memorizing templates: where the reward is expressive, behaviour improves in clinically meaningful ways; where the reward is coarse, limitations remain.


This work has one primary limitation. Our reward relies on CheXbert’s 14 chest X-ray observations, which provides clinically relevant supervision but remains coarse-grained. As a result, it is relatively insensitive to clinically important details that fall outside this label set, such as fine-grained anatomical localization. Future work should therefore explore richer and more granular supervision signals (including calibrated handling of uncertainty and severity), extend the framework to multi-organ, multi-modality and longitudinal settings. In practice, we envision MRG-R1-style models being integrated as decision-support tools for report drafting and quality assurance, flagging potential polarity errors, omissions, or inconsistencies, rather than replacing radiologist judgment, providing a foundation for progressively more reliable and clinically aligned medical report generation systems. 

\FloatBarrier


\section{Conclusion}

This work shows that medical report generation can be improved by optimizing clinical meaning at the report level, rather than relying only on token-level likelihood training. By formulating clinically grounded, report-level rewards and optimizing them with a group-based policy optimization scheme (GRPO), MRG-R1 achieves strong clinical efficacy on IU X-Ray and competitive performance on MIMIC-CXR under a unified CheXbert-based evaluation protocol. 
Beyond the specific model, our key takeaway is a practical recipe for clinically aligned generation: (i) convert clinically meaningful findings into an explicit reward signal, (ii) optimize complete reports with stable group-relative updates, and (iii) apply lightweight structural constraints to produce more consistently organized outputs. 
Limitations include reliance on CheXbert-based labels. Future work will incorporate richer clinical supervision and expert assessment, and extend the framework to broader modalities and settings. 

\section*{References}
\bibliographystyle{IEEEtran}
\bibliography{references} 

\end{document}